# Adopting Explainable-AI to investigate the impact of urban morphology design on energy and environmental performance in dry-arid climates

Pegah Eshraghi[1 2], Riccardo Talami[3 4 *], Arman Nikkhah Dehnavi[1], Maedeh Mirdamadi[1], Zahra-Sadat Zomorodian[1]

[1] *Department of Construction, Shahid Beheshti University (SBU), Tehran, Iran*

*2 Department of Built Environment, College of Design and Engineering, National University of Singapore, 117566, Singapore.*

*3 Department of Architecture, College of Design and Engineering, National University of Singapore, 117566, Singapore.*

*4 Singapore-ETH Centre, Future Resilient Systems II, 1 Create Way, 138602, Singapore.*

## Abstract

In rapidly urbanizing regions, designing climate-responsive urban forms is crucial for sustainable development, especially in dry-arid climates where urban morphology has a significant impact on energy consumption and environmental performance. This study advances urban morphology evaluation by combining Urban Building Energy Modeling (UBEM) with machine learning methods (ML) and Explainable AI techniques, specifically Shapley Additive Explanations (SHAP). Using Tehran's dense urban landscape as a case study, this research assesses and ranks the impact of 30 morphology parameters at the urban block level on key energy metrics (cooling, heating, and lighting demand) and environmental performance (sunlight exposure, photovoltaic generation, and Sky View Factor). Among seven ML algorithms evaluated, the XGBoost model was the most effective predictor, achieving high accuracy ($R^2$:0.92) and a training time of 3.64 seconds. Findings reveal that building shape, window-to-wall ratio, and commercial ratio are the most critical parameters affecting energy efficiency, while the heights and distances of neighboring buildings strongly influence cooling demand and solar access. By evaluating urban blocks with varied densities and configurations, this study offers generalizable insights applicable to other dry-arid regions. Moreover, the integration of UBEM and Explainable AI offers a scalable, data-driven framework for developing climate-responsive urban designs adaptable to high-density environments worldwide.

## 1. Introduction

More than half of the global population currently resides in urban areas, and this figure is projected to reach 68% by 2050 [1]. As cities grow in scale due to rapid urbanization and economic expansion [2][3], urban morphology increasingly influences microclimates and exacerbates challenges like the urban heat island effect [4]. This makes urban form a critical factor in high-density development, drawing attention from architects and urban planners [5]. Urban morphology—encompassing the form, layout, and density of buildings and urban blocks—directly affects shading, solar access, wind patterns, and heat retention, all of which impact energy consumption and environmental performance [6][7]. As a result, urban form is central to sustainable development [8][9].

To create sustainable urban environments, it is essential to integrate energy considerations into urban planning and design [10]. Urban energy planning guides human settlements—from neighborhoods to entire cities—toward a more sustainable future by assessing the potential impacts of design choices on energy consumption [11][12]. Urban Building Energy Modeling (UBEM) enables large-scale energy assessments, accounting for the interactions between buildings and their surroundings [13][14][15]. However, incorporating multiple urban morphology

parameters—such as building height, density, and orientation—into energy planning presents challenges due to the complexity of their interactions [16][13].

In dry-arid climates, incorporating energy considerations into urban planning is crucial due to high temperatures, intense solar radiation, and minimal rainfall. In these regions, inadequate planning can lead to increased temperatures and a greater reliance on HVAC systems for thermal comfort [17]. In Iran, specifically, climate and energy considerations have received limited attention in national urban planning efforts [18]. This study examines the effects of urban morphology on energy consumption and environmental performance, adopting Tehran, Iran, as a representative case. Tehran's range of high-density cores and lower-density areas serves as a useful model for analyzing diverse urban configurations typical of other cities in similar dry-arid climates. This range enables an analysis that encompasses both dense urban patterns and mixed structures, supporting findings applicable to various urban fabrics in dry-arid climates. Previous studies have primarily focused on either broad (national) or small (building) scales, leaving a gap in understanding urban morphology's impact at the urban block level. Moreover, global studies have shown that the effects of urban form on energy metrics are context-specific, making it challenging to generalize findings across different cities [19][17]. Tailored research is needed to explore these relationships in more detail.

This study explores the relationship between urban morphology and energy metrics—specifically cooling, heating, and lighting energy consumption—at the urban block scale, with the findings intended to inform urban planning broadly. In addition to energy metrics, environmental performance metrics such as sunlight hours on façades, photovoltaic (PV) power generation, and Sky View Factor (SVF) are assessed. Through a comprehensive sensitivity analysis, the study examines both typical urban parameters and the heights and distances of surrounding buildings.

A further contribution to knowledge offered by this study lies in its holistic approach, integrating multiple metrics to provide a more comprehensive understanding of how urban form influences microclimates, shading, and energy efficiency. While SVF and solar access are often viewed as intermediate variables, their inclusion here as performance metrics underscores their critical role in shaping microclimates and indirectly influencing energy consumption. Additionally, this study leverages machine learning models (MLMs) and Shapley Additive Explanations (SHAP) to conduct a detailed sensitivity analysis, setting it apart from previous studies [20] that have used more conventional methods. By employing these advanced tools, the analysis achieves a higher degree of transparency and interpretability, allowing for a clearer identification of influential variables. Additionally, by analyzing building height and distance from adjacent buildings as separate indicators, this study offers more granular insights into their specific contributions to energy and environmental outcomes. To guide the research, the following questions are addressed:

1. How do urban morphology parameters—such as building shape, height, and density—affect energy consumption metrics (cooling, heating, lighting) and environmental performance metrics (sunlight hours on façades, PV power generation, and SVF)?
2. Which urban morphology variables exert the most significant influence on these energy and environmental metrics, based on a comprehensive sensitivity analysis?
3. How can this knowledge be utilized by urban planners and policymakers to design more energy-efficient, climate-responsive, and environmentally sustainable urban forms, particularly in dry-arid climates like Tehran?

This study aims to (1) explore the relationship between urban morphology and energy metrics at the urban block scale; (2) conduct a sensitivity analysis to identify the most impactful urban form variables on energy consumption, sunlight hours, PV power generation, and SVF; and (3) provide practical recommendations for urban planners to optimize energy performance and environmental quality across diverse urban configurations in dry-arid climates.

## 2. Literature review

### 2.1. Sensitivity analysis in UBEM

Urban form and morphology play a critical role in building energy performance and environmental outcomes. Numerous studies across different climates have explored these relationships. For instance, Zhang and Gao [21] evaluated the effect of urban form on micro-climate and energy at the block scale in Nanjing, China, finding that energy loads can vary by 23.4% when the microclimate effect is ignored. Similarly, Shang and Hou [3] demonstrated that non-optimal building arrangements can result in a 30% increase in energy consumption for offices and 19% for housing, highlighting the influence of surrounding buildings on energy use. It was then recommended to consider urban form parameters (e.g., building shape, density, green ratio, rotation, street geometry, SVF) at the early design stage to minimize energy demand. Samuelson et al. [22] emphasized the need to account for the urban context in energy simulations, as overlooking surrounding buildings can lead to errors in energy estimates, with generalized contexts producing smaller errors (0-11%) compared to entirely ignoring the context (8-31%). Leng et al. [23] further demonstrated that considering surrounding buildings can lead to a 16-18% reduction in cooling loads in warm, humid climates, underscoring the role of urban morphology in influencing energy consumption. Street geometry and rotation also influence local thermal comfort and energy consumption [24][25]. Yu et al. [26] and Mangan et al. [27] identified building height, floor area ratio (FAR), and building coverage ratio as the most influential factors on energy performance, in temperate-humid climates. Yu et al. [26] found that higher FAR and building coverage ratios significantly impact energy performance in residential communities, with building height and high-rise proportion having a moderate but still important effect.

In arid and semi-arid climates, several studies offer insights relevant to this research. Rode et al. [28] found that compact and tall building typologies provide the greatest heat-energy efficiency at the neighborhood scale, while detached housing yields the lowest efficiency. Natanian et al. [29] identified that in Mediterranean climates, cooling energy intensity decreases as footprint area increases, an important consideration for arid regions where cooling demand is high. Quan et al. [30] revealed that higher building density in Portland, Oregon, led to lower Energy Use Intensity (EUI), which supports similar trends observed in other semi-arid cities. In addition to the reviewed studies, Table 1 lists further relationships observed in the literature. This table focuses on the key highlight from each study that investigates how changes in different inputs affect the outputs.

Table 1. Summary of relationships between urban and building design variables and outputs from further studies in literature.

| Ref. | Climate/ city | Variables | Outputs |
|---|---|---|---|
| [8] Quan et al. | China | Increase coverage ratio | EUI increasing |
| [31] Vartholomaios | Greek Mediterranean | Increase compactness with south rotation and courtyard building typology | Increase energy efficiency |
| [32] Martilli | Hot and dry | Increase compactness, Decrease surface to volume ratio (S/V) | Decrease cooling/ heating demand |
| [23] Leng et al. | Cold region | Increase building coverage ratio, FAR, heights, street ratio, and decrease green space ratio | Heating energy decreasing |
| [3] Shang and Hou | China | Increase heights density and compactness | Micro-climate temperature and energy demand decreasing |
| [33] Natanian et al. | - | Decrease density typology, Increase shape factor | Increase load match |
| [26] Yu et al. | - | -Increase FAR, height, household density, high-rise proportion, - Increase coverage ratio | -Decrease EUI - Increase EUI |
| [34] Natanian and Wortmann | Warm urban areas | Increase PV generation | Increase load match |
| [35] Chen et al. | - | - Increase NDVI - Increase building footprint and building volume | - Decrease EUI - Increase gas and electricity demand |
| [36] Czachura et al. | - | - Increase FAR - Decrease S/V | - Decrease received daylight - Increase compactness, Decrease EUI |
| [37] Saad and Araji | - | - Increase height - Increase height and building form | - Increase EUI, Less sDA - Increase energy generation |
| [38] Pan and Du | Shenzhen, China | Increase density of urban parks, SVF and albedo | Increase visibility and daylight uniformity |
| [4] Xu et al. | Hot and dry, China | - Decrease SVF - Increase height | - Decrease received radiation - Decrease mean radiant temperature |

This study builds on these findings by focusing on the dry-arid climate zone. The selection of studies prioritized those examining the relationship between urban form, energy performance, and climatic context, with a particular focus on energy metrics. The outcomes of these studies present diverse findings, making it challenging for researchers, urban planners, and policymakers to draw consistent conclusions. Specific sensitivity analyses are recommended to better understand the relationships and impacts between urban morphology variables and energy metrics.

In conclusion, while existing literature provides valuable insights, the diverse and context-specific nature of the findings highlights the need for tailored analyses. This study aims to address that need by providing accurate and practical insights for urban energy modeling and planning in Tehran. The goal is to develop energy-efficient and environmentally friendly design strategies at the urban fabric level, considering the unique climatic and urban conditions of the region.

### 2.2. Modeling approaches

A decade ago, city-scale simulations were impossible due to computational and data collection challenges [39]. Recent studies have developed various methods and tools to facilitate modeling and simulation at the urban scale. Current UBEM workflows automatically generate thermal models from 3D building models and archetype attributes, allowing the simulation under specific climatic conditions [10]. Physics-based UBEM models use heat and mass balance equations for spatial and temporal resolution [9]. Swan and Ugursal [40] classified UBEM methodologies into top-down and bottom-up approaches and Kavgic et al. [41] focusing on the latter. In fact, since 2018, bottom-up approaches have gained more attention, leading to new studies [42]. Top-down models use aggregate data to relate energy consumption to factors like socio-economic conditions [1][43]. These models are effective for large-scale analyses but lack granularity, as they treat groups of buildings as units, making them unsuitable for detailed neighborhood-specific energy assessments [9]. Additionally, top-down models often fail to account for the unique characteristics of individual buildings, leading to less accurate energy consumption estimates [44].

In contrast, bottom-up models estimate individual building energy consumption using extensive disaggregated data [10]. They use detailed engineering (white-box), reduced-order (gray-box), or data-driven statistical or MLMs (known as black-box), providing granular analysis but requiring significant data and computational resources [15][45]. Therefore, this study will utilize a bottom-up approach to comprehensively analyze urban energy consumption at the block scale, because it allows for detailed and accurate assessment of individual building characteristics and their interactions with the urban environment.

### 2.3. Using MLMs and SHAP in UBEM

Recent advancements in ML have increasingly facilitated decision-making in UBEM [46]. One notable method is SHAP, which assigns values to features based on their contribution to target predictions [47][48]. SHAP has been employed in various studies to evaluate factors influencing energy consumption [49][50], thermal comfort [51][52], urban heat island [53][54], as well as in the development of tools for building performance analysis [50][55]. For instance, Yu et al. [53] applied SHAP with XGBoost to study the impact of landscape patterns on land surface temperature, finding that 3D landscape metrics were effective in explaining temperature variations. Additionally, Kim et al. [56] identified built-up areas and surrounding vegetation as key factors affecting surface temperature predictions, while Meddage et al. [57] used SHAP to explore the positive and negative effects of geometric parameters on wind pressure in urban settings. Shen and Pan [55] developed a tool for building energy performance assessment using SHAP, Design Builder, and Bayesian optimization LightGBM components [55]. Similarly, Shams Amiri et al. [58] demonstrated the effectiveness of SHAP in forecasting building energy usage and analyzing the impact of urban planning decisions. Given the potential of ML and SHAP in UBEM, this study leverages these tools to investigate how urban morphology impacts energy consumption and environmental performance metrics.

The literature reveals several key trends in UBEM. First, incorporating the urban context—particularly the heights and distances of neighboring buildings—is crucial for accurately assessing environmental performance metrics, even when focusing on individual buildings. Neglecting the impact of surrounding structures can result in significant errors in energy and environmental predictions. While energy performance has been extensively explored in various climates, recent studies highlight the need for more holistic analyses. Metrics such as solar hours received on façades and SVF are increasingly being recognized for their roles in influencing daylight access, microclimate regulation, and outdoor comfort. These factors contribute to a more comprehensive understanding of how urban morphology impacts not only energy use but also broader aspects of livability and environmental sustainability.

Lastly, the literature emphasizes the importance of context-specific analyses, as many metrics are highly dependent on local climate and urban conditions. Although this study focuses on Tehran's dry-arid climate, the methodology and insights can be applied to other regions seeking to optimize urban design for energy and environmental performance. By integrating MLMs and SHAP, this research aims to advance understanding of how urban form influences a range of performance metrics in a more interpretable and comprehensive manner.

## 3. Methodology

The research methodology comprises three steps: (1) Performance Simulation at Urban Level: parametric modeling of various inputs such as urban block length, street width, FAR, building coverage ratio, rotation, buildings height and distances and the simulation of energy consumption metrics, the number of hours of sunlight received by building facades, photovoltaic power generation and SVF as outputs. (2) Building of Predictive Models: multiple predictive models were constructed to forecast the desired output, and the efficacy of each machine learning algorithm was evaluated. (3) Integration of Shapley Values for Feature Importance Assessment: SHAP analysis technique was employed for sensitivity analysis to identify the most influential features, thereby enhancing the understanding of the underlying relationships between the variables and the outputs.

### 3.1. Performance Simulation at Urban Level

Due to the variation in building heights and street widths in the urban context, a series of urban blocks have been developed. These blocks were modeled according to the typical climate and configuration observed in Tehran, Iran (35.6892° N, 51.3890° E). The Rhino environment, Grasshopper, and plugins such as Ladybug, Dragonfly and Honeybee version 1.6.0 have been employed for the modeling and simulation of urban blocks. In the energy modeling section, the primary EPW weather file was modified using the Urban Weather Generator (UWG) plugin to account for local microclimate conditions. This adjustment is crucial since urban climate, influenced by city structure, block texture, building form, and open space arrangement, affects building cooling and heating loads, thermal comfort, and overall performance. Accurate weather data is essential for precise thermal energy demand modeling [10][59]. By incorporating local weather data (e.g., temperature, wind speed, solar radiation), we ensure that the boundary conditions of the urban environment, a key input for building energy modeling, are accurately represented.

The modeling is parametrically conducted on a regular grid, with buildings between 3 and 10 floors, each floor measuring 3.5 meters high. Table 1 lists the modeling variables along with their discretized values. The variables selected for this study were chosen based on their documented significance in influencing energy consumption and environmental performance, as identified in previous research. Key parameters such as site dimension, street width, building typology, FAR, and building coverage ratio directly affect the urban form and spatial arrangement of blocks, which in turn influence shading, solar exposure, and airflow [3][26]. Building dimensions, including width, length, and height, were calculated post-modeling to ensure that their impact on metrics was captured comprehensively. Additionally, the heights and distances of neighboring buildings were treated as separate variables, as research has shown that both factors significantly affect shading and solar access [22][28]. Some parameters, such as WWR, construction materials, and internal loads, were held constant across scenarios to focus on urban morphology variables, as these have been identified as the primary drivers of energy performance at the urban block scale [28]. This approach ensures that the study emphasizes the critical urban form parameters while maintaining a consistent modeling environment for valid comparison across scenarios.

With this framework established, the analysis begins with an examination of two distinct building typologies: rectangular and L-shaped designs. Rectangular buildings are the most prevalent in Iran and were selected to represent common local building fabrics, essential for understanding energy performance. While traditional courtyard houses are also common, they have been well explored in previous studies in Iran [60][61][17]. The less common L-shaped designs were chosen for their unique geometric characteristics, providing insights into optimizing modern architectural forms. The selected building types aim to enhance the overall understanding of diverse building designs within the local context and their implications for energy efficiency.

As detailed in Table 2, the minimum area for an urban block is 1 hectare (100 meters by 100 meters). Street widths follow the specifications of Tehran's master urban plan, with internal block access streets measuring 6 and 12 meters, and main streets set at 20 meters. The building coverage ratios of 60% and 45% and the WWR of 35% and 45% for residential and commercial buildings, respectively. The urban block rotations of 40, -40, and 0 degrees relative to north were selected to reflect the typical orientations found in Tehran's existing urban layout. By defining the number of divisions along the X and Y axes, the minimum of 6 and maximum of 24 parcels in urban blocks are generated.

Additionally, the green space ratio is designated as 20% and 40% of the site area. In this context, the principal activity zone has been as a minimum occupancy coefficient of 50% and 70%. It is important to highlight that these activity zones represent urban areas where the primary focus is on residential use, but they also accommodate commercial and office services.

Table 2. Modeling variables, discrete values and units of measurements.

| Variable | Discretized values | Unit |
|---|---|---|
| Site length | 100, 120, 140 | m |
| Street width | 6, 12 | |
| Building typology | Rectangular, L-shaped | - |
| FAR | 4.5, 6.5 | |
| Rotation | -40, 0, 40 | degree |
| Building coverage ratio | 45, 60 | |
| Green space ratio | 20, 40 | % |
| Residential ratio | 50, 70 | |

The three building uses—residential, office, and retail—were modelled according to their respective archetypes and their construction material and internal heat loads values were derived from Iran national regulations [62] (Table 3).

Table 3. Fixed design and operational values of the residential, office and commercial archetypes.

| Fixed values | | Residential | Office | Commercial | Unit |
|---|---|---|---|---|---|
| Plug/equipment load | | 4 | 14 | 4 | $W/m^2$ |
| Infiltration rate | | | 3 | | ACH |
| R-value | Roof | | 1.8 | | $m^2K/W$ |
| | Wall | | 0.9 | | |
| | Ground | | 0.5 | | |
| Window U-value | | | 2.7 | | $W/m^2K$ |
| Lighting density | | 10 | 11.5 | 16.9 | $W/m^2$ |
| HVAC system | | | Ideal air load | | - |
| WWR | | 35 | 40 | 50 | % |
| Visual transmittance | | | 0.75 | | - |
| SHGC | | | 60 | | % |
| Anthropogenic heat | | | 14 | | $W/m^2$ |

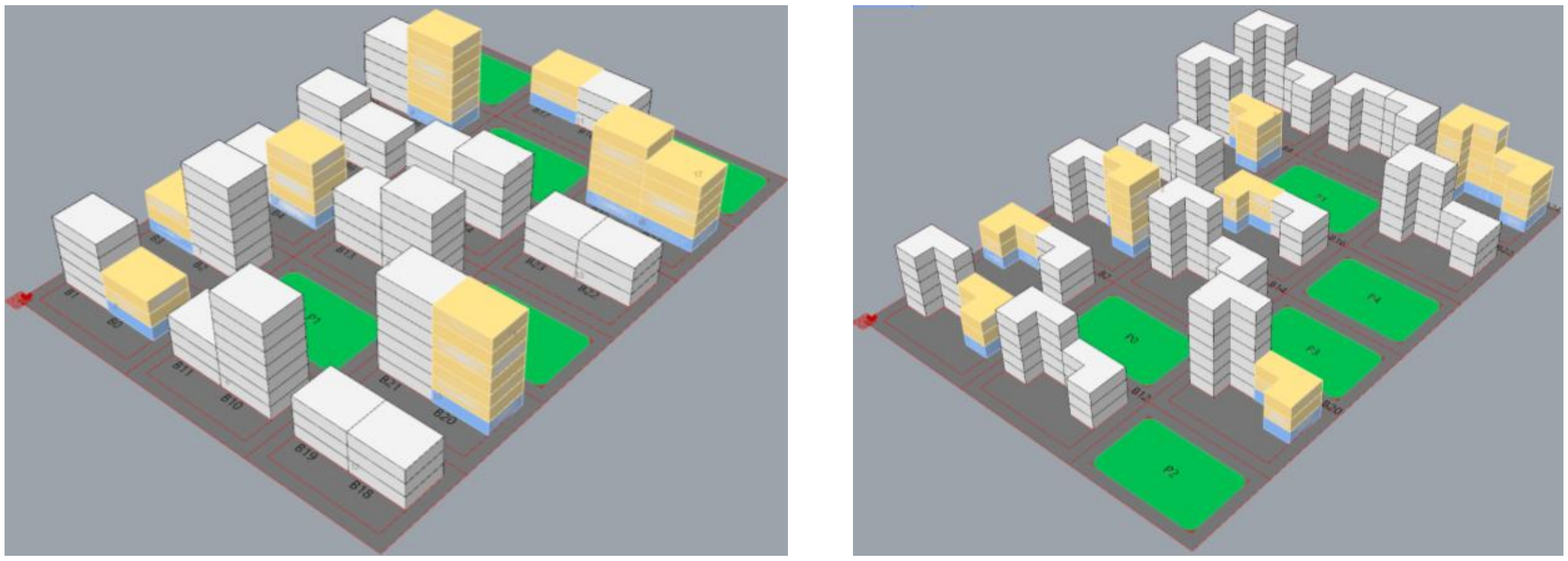

Figure 1. Exemplary modelled urban layout for both rectangular and L shape forms.

The investigated output metrics are annual energy consumption (cooling, heating, lighting), sunlight hours received on facades, PV power generation and SVF. Different engines can be used to calculate energy metrics in Grasshopper, such as OpenStudio and UrbanOPT [63]. Prior to the modelling step, UrbanOPT and OpenStudio were compared to determine which software provides more efficient and accurate energy metric calculations. UrbanOPT, used with the Dragonfly plugin, operates with the geoJSON format. This allows for direct import into EnergyPlus, which is efficient for network calculations and urban infrastructure simulations. On the other hand,

OpenStudio, used with the Honeybee plugin, also leverages EnergyPlus but requires converting the Dragonfly model into a Honeybee-compatible format. OpenStudio appears to be better suited for detailed building-level calculations. Although both engines produced similar results in terms of accuracy, OpenStudio was chosen due to a faster simulation execution time. In addition, the energy modeling approach utilized in this study involves simulating ground, intermediate, and top floors. For intermediate floors with the same archetypes and assumptions, results were aggregated through multiplication.

In this study, not only are energy consumption metrics such as cooling, heating, and lighting energy analyzed, but intermediate parameters such as SVF and sunlight hours received on façades are also included. These intermediate parameters are considered essential for providing insights into the broader environmental impacts of urban morphology on microclimate and livability, which are not fully captured by energy consumption metrics alone. While SVF may be regarded as an intermediate metric, its role in influencing microclimatic conditions, such as heat dissipation and wind patterns, is seen as critical for ensuring a comprehensive assessment of urban sustainability. Furthermore, understanding the relationship between urban form and SVF is necessary for optimizing both energy efficiency and environmental quality in high-density areas. By incorporating these parameters, the analysis is intended to provide a more holistic framework that extends beyond energy performance to include factors contributing to the overall livability and environmental sustainability of urban areas.

SVF represents the fraction of visible sky [64] and is computed using the Ladybug view percent component. This component emits multiple rays from the observation point towards the sky, determining the proportion of rays that successfully reach the sky without encountering obstacles. In this study, the buildings' yards parcel is divided into 1 x 1 meter grids, and the SVF is calculated at each grid point. The average SVF for the entire surface is then computed from these values. The PV power generation output has been calculated using the Ladybug Photovoltaics Surface component. The assumptions are that the solar panels are installed on 40 percent of the roof, oriented south at a 32-degree angle, with an 18 percent efficiency of the silicon crystal type. Finally, the sunlight hours received on facades output for each building's entire envelope, divided into 1 x 1 meter grids, is also calculated using the LB Direct Sun Hours component.

In addition to the mentioned outputs, the following metrics are also recorded: (1)Distances and Heights of Surrounding Buildings: The distance and height of adjacent buildings are measured in eight cardinal and intercardinal directions (N, NE, E, SE, S, SW, W, and NW), (2)Height of Adjacent Buildings: The heights of buildings adjacent to each block are recorded, providing additional context for evaluating urban shading and building massing, and (3)Dimensional Properties: The length, width, and number of stories of the buildings generated in each iteration are documented. This detailed recording ensures a comprehensive understanding of how the spatial arrangement of neighboring buildings influences the energy performance and environmental metrics of the modeled urban blocks.

In this study, the height and distance of surrounding buildings were analyzed as separate variables to isolate their distinct impacts on energy consumption, PV power generation, sunlight hours received on façades, and SVF. This approach allows for a more precise understanding of how each factor independently influences energy and environmental performance. While it is acknowledged that these factors could interact, separating them in the analysis provides valuable insights into their specific contributions, which may be overshadowed if analyzed together. By treating them individually, this study highlights the unique role that both height and distance play in shaping urban microclimates, enabling more targeted recommendations for urban planning and design.

Over the course of 5 months, using 3 external servers, 2,400 urban blocks were simulated, comprising 48,000 building and park parcels. Finally, Colibri and TT Toolbox plugins generated the parametric dataset from the specified inputs and outputs, with the results stored in .csv format.

### 3.2. Building prediction models

After pre-processing the data, a total of 36,060 parcels—comprising 29,560 buildings and 6,500 parks—were retained for creating the MLMs. The models were trained using a set of key variables, including building length, width, height, coverage ratio, height and distances of adjacent neighboring buildings, green space ratio, street width, density, rotation, building function, and shape. This paper explores various machine learning algorithms, such as Catboost, XGBoost, LightGBM, Random Forest (RF), Artificial Neural Networks (ANN), K-Nearest Neighbors (KNN), and Support Vector Machines (SVM). These algorithms were chosen for their ability to effectively handle different types of datasets and generate accurate predictions across a wide array of domains and applications. This paper aims to achieve robust predictive capabilities and comprehensive insights for the subsequent analyses by exploring various algorithms, each algorithm offers unique strengths and capabilities, enabling a more thorough exploration and understanding of complex data patterns. This strategy increases the confidence and prediction in the extraction of meaningful insights from the dataset. Catboost [65], optimized for categorical data; XGBoost [66], known for high performance and speed in gradient boosting; LightGBM [67], efficient with large datasets and light on memory usage; RF [16], combining multiple decision trees for robust predictions; ANN [68], mimicking the human brain to recognize patterns and relationships; KNN [68], classifying data points based on proximity to others; and SVM [26], finding the optimal hyperplane to classify data. Initially, default hyperparameters were selected train the models. Subsequently, the performance of each model was evaluated to determine whether hyperparameter tuning was necessary, as this process can significantly enhance model accuracy and effectiveness by optimizing the parameters for the specific dataset and task. Table 4 breakdowns the tuned MLM hyperparameters utilized in this study. This tuned set of hyperparameters ensures the robustness and effectiveness of the CatBoost algorithm in capturing complex patterns within the data while generalizing well to unseen instances.

Table 4. Tuned MLMs hyperparameters.

| Algorithm | Hyperparameters |
| --- | --- |
| CatBoost | Learning Rate: 0.1, Depth: 6, L2 Regularization: 3, Loss Function: RMSE, Number of Trees: 100, Random Seed: 42, Early Stopping Round: 10 |
| XGBoost | Learning Rate: 0.1, Max Depth: 6, Min Child Weight: 1, Gamma: 0, Subsample: 1, Colsample Bytree: 1, Num Round: 100 |
| LightGBM | Learning Rate: 0.1, Max Depth: 6, Min Child Weight: 1, Subsample: 1, Colsample Bytree: 1, Num Leaves: 31, Num Iterations: 100 |
| RF | Number of Estimators: 100, Maximum Depth: None, Minimum Samples Leaf: 1, Minimum Samples Split: 2 |
| ANN | Hidden Layers: (100, 50), Activation: ReLU, Solver: Adam, Learning Rate: 0.001, Batch Size: 32, Epochs: 100 |
| KNN | Number of Neighbors: 5, Weight Function: Uniform, Distance Metric: Euclidean |
| SVM | Kernel: RBF, C (Regularization Parameter): 1.0, Gamma: Scale |

## Models' evaluation

To assess the performance of models, specific evaluation metrics are applied. For outputs derived from regression models, $R^2$ is employed as one of the standard metrics for evaluating the quality of these models. This metric represents the proportion of the variance in the dependent variable that is explained by the independent variables within the model. A value of $R^2$ closer to 1 indicates a better fit of the model to the data. The formula for calculating $R^2$ is as follows (Equation 1):

$$R^2 = \frac{SSR}{SST} - 1 \tag{1}$$

where:

- SSR (Sum of Squares of Residuals): The sum of the squared differences between the observed values and the values predicted by the model.
- SST (Total Sum of Squares): The sum of the squared differences between the observed values and the mean of the observed values.

This index alone may not adequately represent the overall quality of a model, particularly in terms of its ability to accurately predict new data. Therefore, it is advisable to utilize additional metrics, such as the Root Mean Square Error (RMSE), to assess predictive accuracy [69][70]. RMSE quantifies the closeness of the predicted values to the actual values (Equation 2). A lower RMSE indicates superior model performance.

$$RMSE = \sqrt{\frac{1}{N}\sum_{i=1}^{N}\left(\hat{Y}_i - Y_i\right)^2} \qquad (2)$$

where:

- N: The number of data.
- $\hat{Y}_i$ :The predicted value.
- $Y_i$ :The observed (or simulated) value.

In addition to $R^2$ and RMSE, another metric used to evaluate model performance is training time. This metric reflects the computational efficiency of each algorithm, with a lower training time indicating faster model training.

### 3.3. Sensitivity analysis

The complexity of MLMs, often perceived as "black boxes," poses significant challenges for domain experts during the decision-making process [71]. The lack of transparency in these models can undermine users' ability to trust and understand the results [72]. To mitigate this issue, post hoc explanation techniques, such as SHAP, are employed to provide interpretable insights. SHAP is a model-agnostic method that explains individual predictions by estimating feature contributions. It calculates Shapley values to assess features interactions, offering a comprehensive perspective on their significance in influencing target variables [46]. SHAP is preferred over other techniques, such as gain and split count, due to its fairness, stability, and ability to explain individual predictions based on game theory, assigning an importance value to each feature and delineating its contribution to the predicted outcome [73]. The SHAP method relies on the following definitions to calculate feature importance (Equation 3) [74]:

$$\phi_i(p) = \sum_{S \subseteq N/\mathrm{i}} (\frac{|S|!(n-|S|-1)^i}{n!}\ [p\,(S \cup \mathrm{i}) - p(S)) \tag{3}$$

- N: A set containing n features
- S: A subset of features that does not include the feature for which $\phi_i$ is calculated
- S∪{i}: A subset of features that includes the features in S plus the feature i
- S⊆N\{i}: All subsets S that are subsets of the complete set N excluding i

The SHAP values for each variable were calculated to quantify their contribution to the relevant metrics. To allow for meaningful comparisons, the SHAP values were normalized between 0 and 100. The normalization was performed separately for each category of variables to avoid any skewing of results by variables with disproportionately large SHAP values. This approach ensures that the influence of each variable within its respective category is accurately reflected, allowing for a clearer comparison between variables. The decision to normalize each category individually, rather than across all variables, was based on the observation that normalizing all variables together would result in unrealistically low percentages for many variables, distorting their relative importance. To enhance clarity and focus on the most impactful factors, the variables are categorized into three distinct groups based on their influence on the metrics, with percentage thresholds defining each category:

- **Dominant Variables**: These variables have the most significant and consistent impact across different scenarios, driving changes in the metrics by more than 50%. These factors play a critical role in determining the overall performance and must be prioritized in the analysis.
- **Influential Variables**: Variables in this category exhibit a moderate impact on the metrics, contributing changes between 20% and 50%. Their influence may vary depending on specific conditions, and while they are important in certain contexts, they do not consistently dominate the results.
- **Negligible Variables**: These variables have minimal or inconsistent effects on the metrics, contributing less than 20% to changes in outcomes. While they may occasionally have localized impacts, they do not significantly alter the overall results.

This categorization is specifically designed to guide decision-making in urban morphology planning, particularly with respect to energy and environmental objectives. By identifying and prioritizing the variables based on their impact, decision-makers can optimize their strategies. For instance, negligible variables can be deprioritized, allowing for a more efficient focus on those factors that significantly influence outcomes.

In addition, to accurately evaluate the results of simulations and analyses, statistical tests are employed to determine whether the observed differences are attributable to random chance or represent genuine effects. One of the key metrics in these evaluations is the P-Value, which is used to assess the statistical significance of the findings. The P-Value quantifies the probability of observing the obtained results under the null hypothesis, which posits that no significant difference exists between groups. By calculating the P-Value using analysis of variance (ANOVA), we can rigorously evaluate whether the differences observed in the data are statistically meaningful. If the P-Value is less than 0.05, it indicates that the observed differences between groups are statistically significant, leading to the rejection of the null hypothesis. This method ensures that the conclusions drawn from the analysis are robust and not influenced by random variability.

# 4. Results

This section presents an evaluation of the predictive accuracy of MLMs used to estimate key energy and environmental metrics, followed by an analysis of urban morphology parameters' influence on these metrics. First, the predictive performance of the MLMs is assessed for accuracy and reliability. Then, using SHAP, the impact of 30 urban morphology parameters is analyzed across six output metrics—cooling, heating, lighting energy demand, PV power generation, sunlight hours on façades, and SVF—to offer insights into optimizing energy efficiency and environmental performance in urban blocks.

## 4.1. Performance of the predictive models:

This section evaluates various MLMs to determine their accuracy in predicting various energy and environmental metrics (Figures 2 and 3). The evaluation of predictive models is critical for gauging their effectiveness in capturing underlying data patterns. The $R^2$ serves as a key metric to quantify the proportion of variance in the output variable explored by the model. Higher $R^2$ values indicate a stronger predictive performance and a closer alignment between the predicted and observed values.

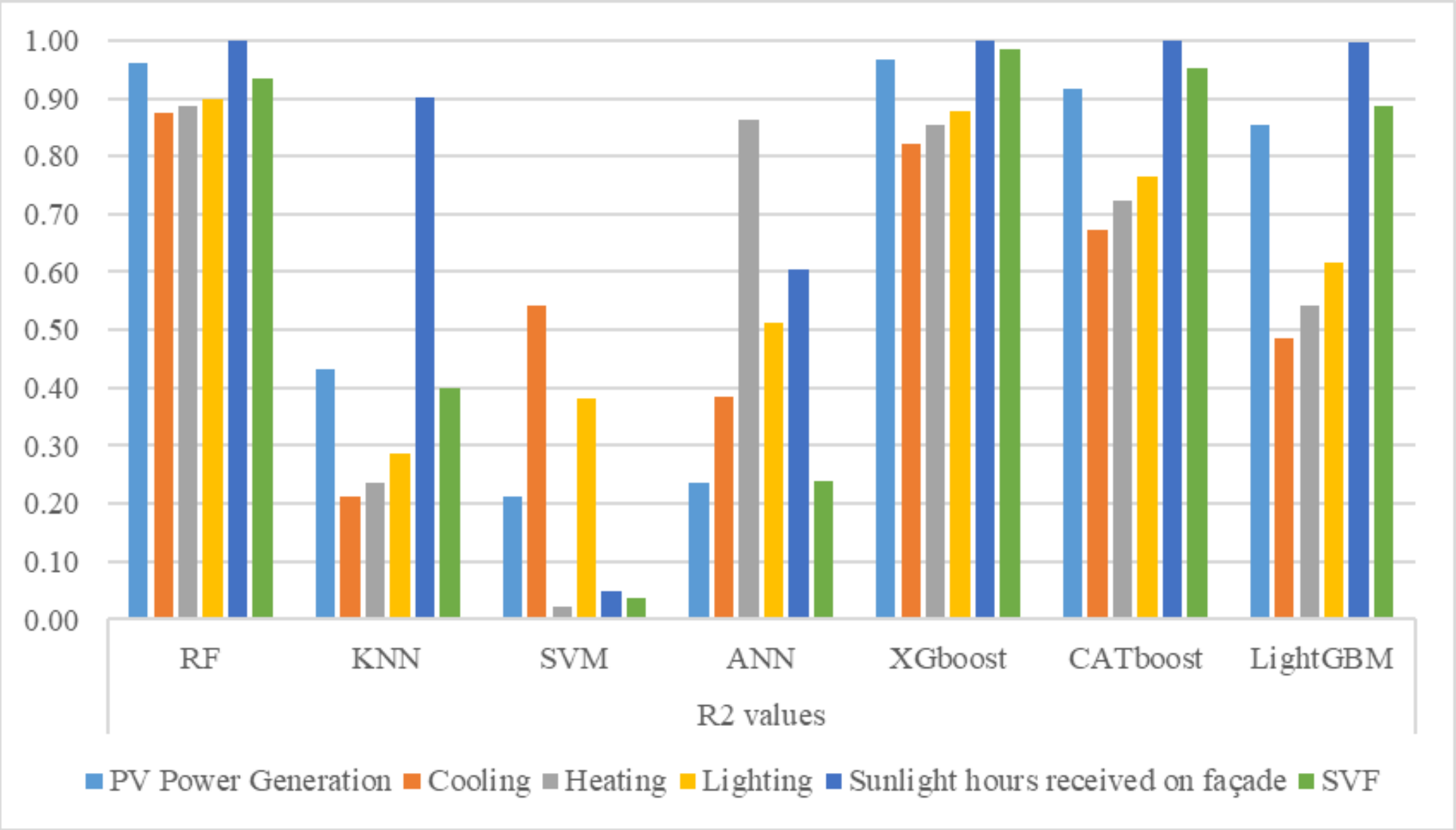

Figure 2. $R^2$ values for different MLMs in predicting different energy and environmental metrics.

In the prediction of PV power generation, RF achieves an $R^2$ value of 0.969, indicating that approximately 0.96 of the variances in PV power output is captured by the model. Similarly, RF exhibits $R^2$ values of 0.87 for cooling, 0.89 for heating, and 0.9 for lighting energy consumption. While other models, such as CatBoost, XGBoost, and LightGBM, also demonstrate strong performance, they generally exhibit slightly lower $R^2$ values across these outputs. For instance, XGBoost achieves an $R^2$ value of 0.97 for PV power generation, slightly higher than the value obtained by RF, but it records $R^2$ values of 0.82 for cooling, 0.85 for heating, and 0.88 for lighting energy consumption. CATboost and LightGBM models exhibit $R^2$ values slightly lower than those of RF and XGboost in forecasting cooling and heating demand, and lighting consumption. In contrast, ANN, KNN, and SVM consistently display lower predictive performance across the majority of outputs.For instance, ANN, KNN, and SVM achieve $R^2$ values of only 0.23, 0.43, and 0.21, respectively, for PV power generation, highlighting their limited ability to capture the complex relationships in this output. Additionally, the RF, XGboost, CATboost and LightGBM models achieve $R^2$ values of 1.0 when predicting sunlight hours received on facades, and $R^2$ values higher than 0.9 for SVF.

Figure 3 presents a heat map reporting the RMSE values. Each cell in the grid represents the intersection of an algorithm and a metric, with the colors indicating the RMSE value. A lighter color signifies better performance of the algorithm in predicting the respective metric. ANN is removed from the figure due to its significantly high values, that would hinder readability.

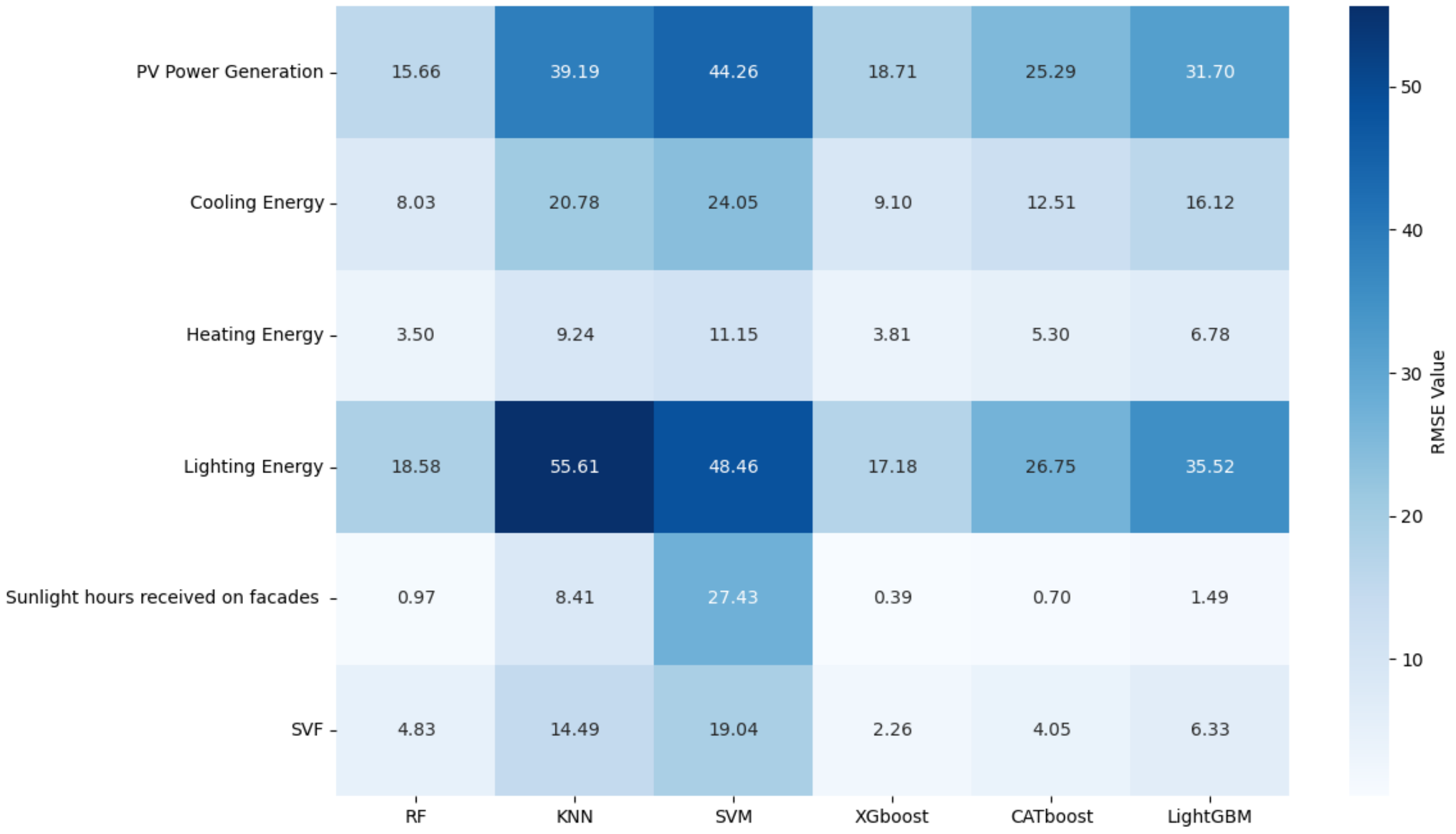

Figure 3. Heatmap of RMSE values between each MLM and energy and environmental metric.

The ANN algorithm consistently yields the highest RMSE values across all outputs. Following, the SVM and KNN algorithms demonstrate high RMSE values of 48.46 and 55.61, respectively, in predicting the lighting energy metric. The LightGBM algorithm consistently exhibits higher RMSE values across outputs compared to other models. In contrast, the RF algorithm shows lower RMSE values for predicting energy metrics compared to XGBoost and CATBoost. However, for metrics such as sunlight hours received on facades, SVF, and PV power generation metrics, the XGBoost algorithm outperforms both RF and CATBoost.

In summary, while the models exhibit varying levels of predictive accuracy across different outputs, RF demonstrates superior performance, as indicated by its high $R^2$ values across all outputs. However, in terms of RMSE, the XGBoost algorithm shows generally lower RMSE values. Therefore, to provide a comprehensive evaluation and aid the decision-making, this paper examines the model training times. Figure 4 displays the average $R^2$ value of each model along with their respective training times.

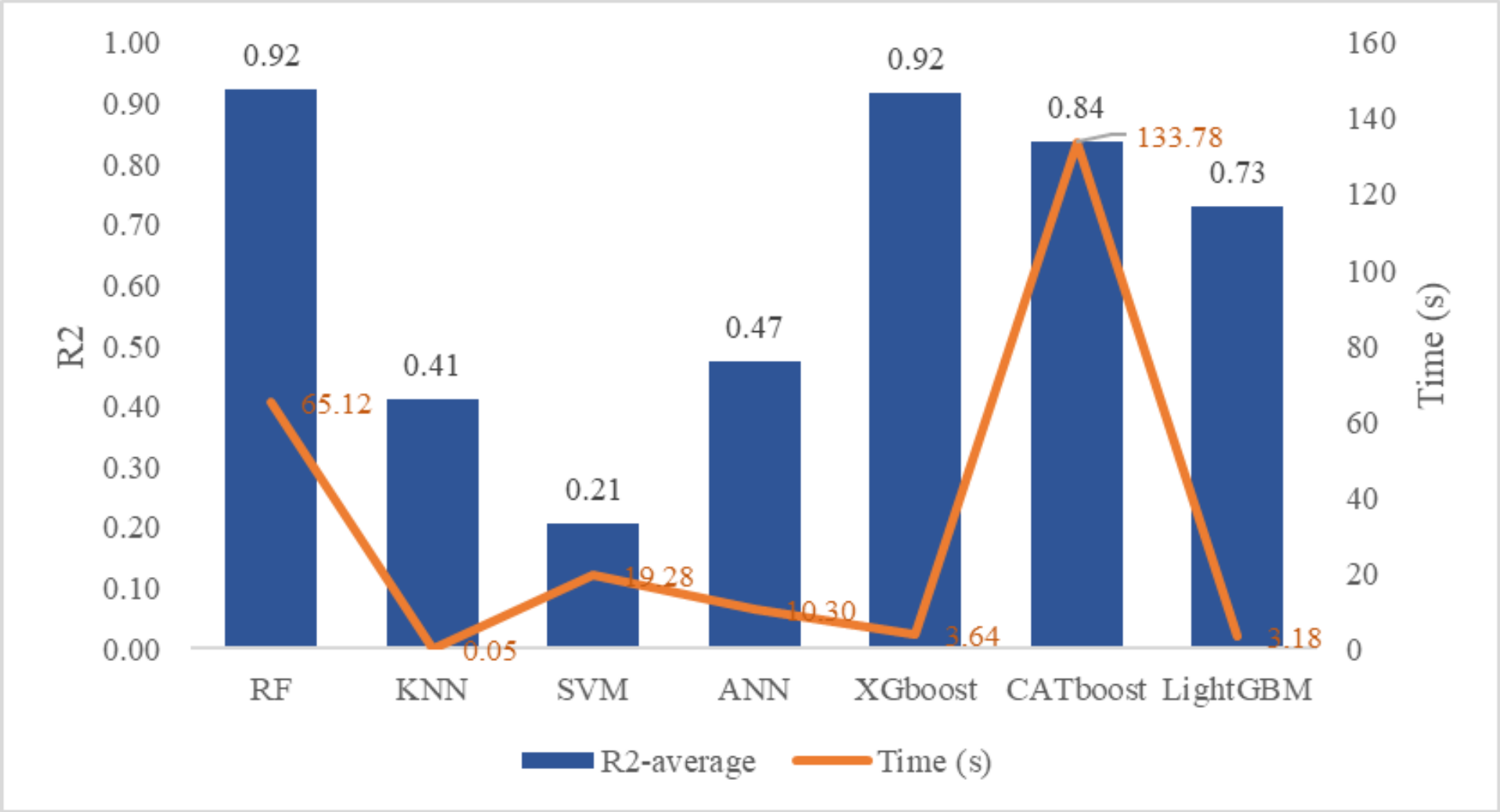

Figure 4. Average $R^2$ value for each MLM along with the time required for training.

The CATBoost model requires the longest training time at 133.78 seconds, while the KNN model has the shortest time at 0.05 seconds. Notably, the XGBoost model achieve accuracy comparable to the RF model with a training time of just 3.64 seconds, compared to65.12 seconds for RF. As a result, the XGBoost model has been employed in this paper for the subsequent analyses.

## 4.2. SHAP analysis results

Figure 5 presents a comprehensive sensitivity and statistical significance analysis, illustrating the magnitude of the relative impact of variables on each metric. The variable with the largest influence on each metric is displayed in the darkest color (dark blue). Variables are ranked according to their cumulative ranks across all metrics, with those having the highest total ranks placed at the top. Variables marked with N/A are influential in the sensitivity analysis but did not demonstrate statistical significance.

Figure 5 provides an analysis of the relationships between 30 input variables and 6 output metrics, illustrating the direction and nature of influence each variable has on these metrics. The relationships are represented using arrows to indicate the type of influence: direct (↑), inverse (↓), or complex (↕). A direct relationship (↑) suggests that increasing/decreasing the variable leads to an increase/decrease in the metric, while an inverse relationship (↓) implies that increasing/decreasing the variable causes a decrease/increase in the metric. In some cases, the relationship between a variable and a metric is complex in both directions (whether the variable increases or decreases), represented by a ↕ arrow, meaning it can vary in a non-linear or unpredictable manner. However, in other cases, the relationship is complex in only one direction: the effect is clear when the variable changes in one direction but becomes complex when it changes in the opposite direction. To distinguish these one-way complex relationships, the table uses colored borders: Red borders indicate that increasing the variable has a clear influence (either direct or inverse), but decreasing it results in a complex relationship. Yellow borders indicate that decreasing the variable has a clear influence, but increasing it results in a complex relationship.

This color-coding helps to clarify which variables exhibit predictable effects in one direction while having complex, less predictable effects in the opposite direction, providing a more detailed understanding of the relationships between the variables and the metrics. For instance, building length consistently shows a direct relationship (↑) with the number of hours of received sunlight. However, when increasing this variable, it also exhibits a direct relationship (↑) with heating energy. On the other hand, decreasing the building length has a complex relationship (↕) with heating energy.

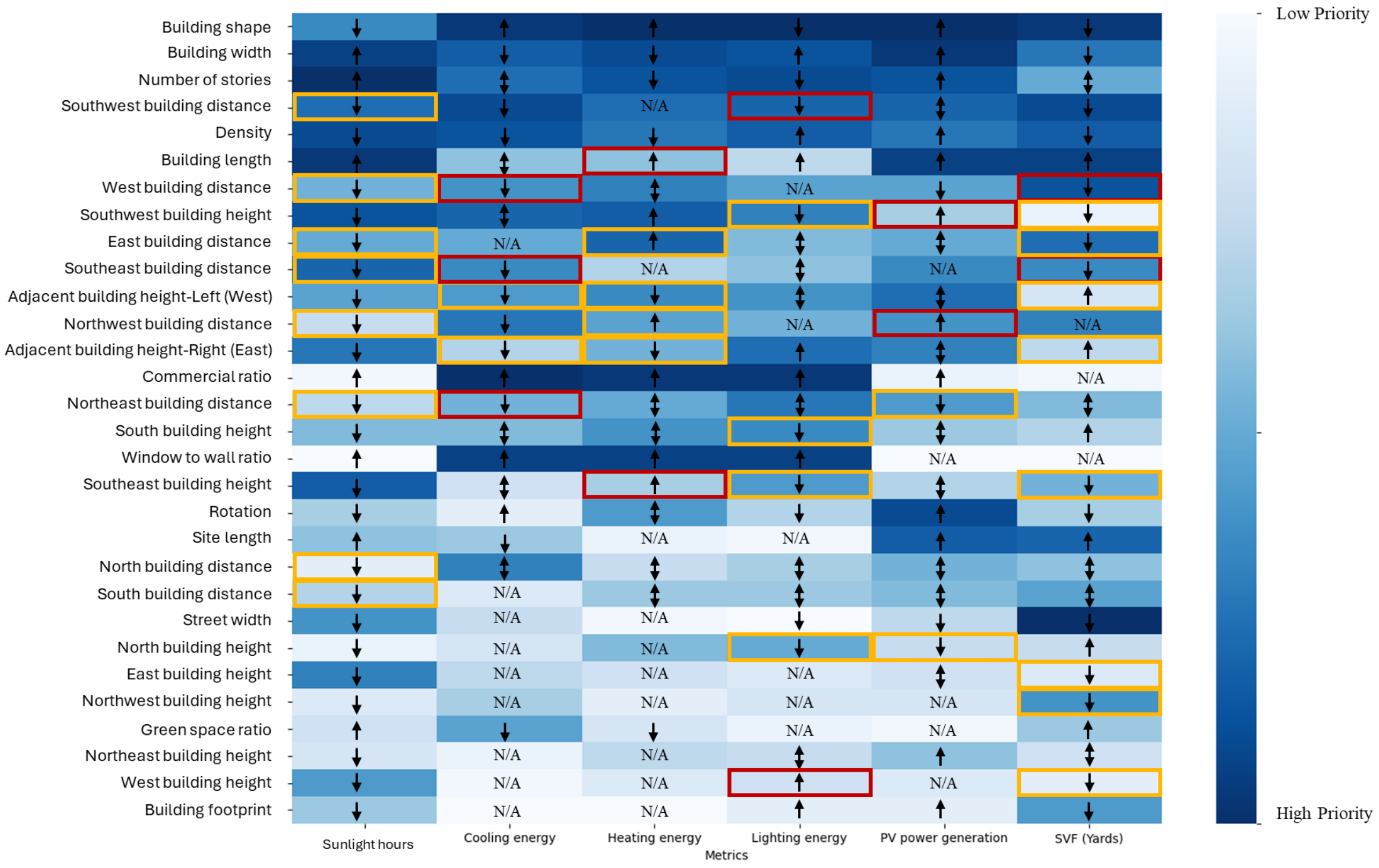

Type of relationship: ↑: direct, ↓: inverse, ↕: complex- Clear effect when decreasing the variable; complex relationship when increasing- Clear effect when increasing the variable; complex relationship when decreasing.

Figure 5. The magnitude and the direction of the impact of variables on the output metrics.

#### 4.2.1. Sunlight hours received on facades

The number of stories has the most significant impact as it directly increases the surface area of the building envelope exposed to sunlight. Building length follows with an influence of 73.5%, as longer buildings provide a larger facade surface, increasing sunlight exposure. Building area, with a 66.35% effect, also plays a crucial role since larger building envelopes capture more sunlight. Building width, contributing 50.8%, affects sunlight hours similarly by expanding the building's exposed surface area. As these values increase, the number of hours of sunlight the building receives also increases.

In the category of influential variables, the density of buildings within the urban plot displays an inverse relationship with sunlight hours, showing a 45.36% effect. Higher density reduces surface exposure to sunlight as buildings become more crowded. The heights of neighboring buildings, particularly in the southwest and southeast, have a moderate influence, with 24.58% and 24.01% effects respectively. These exhibit an inverse relationship with the metric, indicating that taller buildings in these positions provide shading, thereby reducing sunlight exposure. Proximity to adjacent buildings also plays a role; the distances from buildings in the southeast and southwest contribute 23.06% and 21.7%, respectively, to the variation in sunlight hours, reducing these distances increases the sunlight hours on facades. A reduction in street width, with a 21.3% effect, can lead to an increase in sunlight hours received on facades, which can be attributed to the modeling approach: narrower streets result in larger plot areas, thus expanding the building envelope area. Similarly, the height of the right (eastern) adjacent building impacts sunlight, with a 20.18% effect, as taller neighboring structures in this position also cast shadows and reduce sunlight hours.

Negligible variables, with less than 20% influence, include factors such as the height of western adjacent buildings and the distance from eastern buildings. These variables have minimal or inconsistent effects on sunlight hours. Additionally, factors like building usage, the height and distance of northern, northwest, and northeast neighboring buildings, as well as the shape of the building (with rectangular buildings receiving more sunlight than L-shaped ones), are categorized as having negligible impact on the sunlight received by facades.

#### 4.2.2. Cooling energy demand

Commercial ratio is the most dominant factor affecting cooling energy consumption. Commercial buildings generally consume more energy for cooling due to higher occupancy levels and different usage patterns compared to office and residential buildings. The shape of the building, with an impact of 94.1%, also plays a crucial role, as L-shaped buildings have a larger surface area exposed to sunlight, leading to increased heat gain and cooling demand. Another critical factor is WWR, with an 80.73% effect, is another critical factor, where larger windows permit more sunlight to enter, thus increasing the cooling energy required to maintain comfortable indoor temperatures.

Several variables exhibit moderate effects on cooling energy consumption. The southwest building distance, with an influence of 42.18%, is particularly important; reducing this distance increases cooling energy due to increased exposure to the afternoon sun. The density of buildings, with a 35.83% effect, generally has an inverse relationship with cooling energy, as higher density lowers the surface area-to-volume ratio, reducing heat gain. Similarly, building width (29.25%) and building area (27.89%) show moderate effects, where larger building dimensions reduce heat gain and cooling energy needs. The green space ratio, with a 25.9% impact, is another key factor, as increased green space can lower ambient temperatures through shading and evapotranspiration, reducing cooling energy demand. Some neighborhood characteristics, such as the height of the southwest neighboring building, with a 21.77% influence, an inverse relationship with cooling energy, as taller buildings can provide shading that reduces cooling needs. The number of stories (21.54%) also contributes to cooling energy demand by increasing the surface area exposed to solar radiation.

The following variables have a negligible impact on cooling energy demand, as their effects are less than 20%. These include the northwest building distance and north building distance, both with a 19.73% influence, as well as the distances from the southeast at 17.91% and west at 17.01%. While proximity to nearby structures can influence cooling demand to a small extent, these factors do not significantly affect the overall energy consumption due to the minimal shading and solar exposure changes they cause. Additionally, site length (10.43%), building length (10.43%), and the height of the southeast building, along with the height of adjacent buildings in the east. These variables have minimal effects on cooling energy consumption. Although the southwest building distance has a somewhat greater influence, variations in the height of adjacent buildings, particularly in the south and northwest, are less impactful. This indicates that proximity to structures and open spaces plays a more important role in cooling energy demand than building height.

In conclusion, changes in adjacent distances, such as dSW, have a larger impact on cooling energy compared to variations in building height, suggesting that proximity to other structures and open spaces is more important than vertical dimensions.

#### 4.2.3. Heating energy demand

Building shape is the most critical factor. Rectangular buildings generally require less heating energy than L-shaped buildings due to their compact form, which minimizes heat loss. The number of commercial floors, with an 81.27% influence, also plays a significant role, as commercial buildings typically require more heating due to their larger spaces and higher energy demands. The WWR, with a 74.6% effect, shows a direct relationship with heating energy demand, as larger windows allow more heat loss. Additionally, building area contributes 51.75%, indicating that larger buildings tend to lose more heat and require additional energy for heating.

Several variables are influential, showing a moderate impact on heating energy demand. Building width, with an influence of 36.19%, exhibits an inverse relationship with heating demand—wider buildings retain more heat, thus reducing the need for heating. Density also demonstrates an inverse relationship, contributing 27.87%, as denser building configurations trap more heat, lowering energy consumption. The height of the southwest building has an impact of 26%, where taller buildings can obstruct sunlight, increasing heating demand. Additionally, the distance to the eastern building, with a 25.3% effect, shows an inverse relationship, where reduced distance decreases exposure to cold winds and lowers heating energy needs.

Negligible variables, with less than 20% influence, include the west building distance (15.24%) and south building height (14.92%), both of which exhibit complex relationships with heating energy consumption, influenced by interactions with other environmental factors. The height of the adjacent building to the left (west), with an effect of 14.92%, is directly related to heating energy consumption, as reducing the height of this building increases heating needs due to diminished solar gain. The south building height, with a 14.92% influence, also exhibits a complex relationship, where its effect on heating demand depends on the interactions with sunlight and shading. Similarly, building rotation affects heating energy consumption in a complex way, impacting the heat gain and loss depending on the orientation of the building. Furthermore, the northwest building distance, with a 14.6% influence, shows an inverse relationship with heating energy demand, where reducing this distance decreases heating needs due to improved thermal insulation from adjacent structures. The northeast building distance, at 13.65%, exhibits a more complex relationship.

#### 4.2.4. Lighting energy demand

The most dominant variables influencing lighting energy demand are building shape and the number of commercial floors, with an 85.71% effect. Rectangular buildings typically require more lighting energy compared to L-shaped buildings due to reduced natural light penetration in deeper spaces, increasing reliance on artificial lighting. The WWR also plays a critical role, with a 77.71% influence, where larger windows allow more natural light to enter, reducing lighting energy consumption. The number of stories, with a 67.43% effect, further contributes to lighting energy demand, as taller buildings may face challenges in distributing natural light evenly across all floors. The number of commercial floors, at 47.4%, directly impacts lighting energy due to the typically commercial spaces requiring more artificial lighting.

Several influential variables also contribute to lighting energy demand. Building area (38.5%), building width (31.14%), and density (30.57%) play a moderate role, where larger and denser buildings tend to have uneven natural light distribution, leading to greater reliance on artificial lighting. The southwest building distance, with an impact of 27.7%, influences lighting energy needs, as increasing the distance to the southwest building allows more natural light into the building. Conversely, the height of the right (East) adjacent building, at 26.57%, tends to increase lighting energy consumption due to potential obstruction of natural light. Additionally, the northeast building distance, as well as the heights of the southwest and south buildings, each with a 23.14% impact, exhibit complex influences; however, reducing the height of these neighboring structures generally allows more natural light to penetrate, thereby decreasing lighting energy needs. The southeast building height and the left (west) adjacent building height, both at 20.86%, also affect lighting energy demand, as lower building heights in these positions allow more natural light to enter.

Lastly, several variables are considered negligible. These include the distance to the west and northwest buildings, site length, green space ratio, and the heights of the east and northwest buildings, which were found to have minimal impact on lighting energy demand. Additionally, street width, building footprint, and the heights of the west and northeast buildings, along with building length, were determined to have the least influence on lighting energy consumption and can therefore be deprioritized in further analysis.

#### 4.2.5. Energy generated by Photovoltaic panels

The dominant variables influencing PV energy generation are building area and building shape, contributing 84.83%. Rectangular buildings typically generate less PV energy compared to L-shaped buildings due to their smaller roof area. In all scenarios, 40% of the available roof area is allocated for PV panels, but L-shaped buildings, with their larger roof areas, can support more PV installations. Building width, with an influence of 68.25%, directly impacts the roof area available for PV panels, while building length (58.5%) and rotation (56.83%) also show a direct relationship, where both variables contribute to optimizing the surface for PV installation. Southward and eastward orientations also generate more PV energy compared to westward orientations. Southward orientation is optimal for capturing sunlight throughout the day, while eastward rotation enhances morning sunlight exposure. In contrast, westward orientations typically receive less consistent sunlight due to late-day shading. The number of stories, with a 57.16% influence, has a direct effect on PV power generation, as taller buildings tend to capture more roof radiation due to reduced shading from neighboring structures.

Several influential variables pertain to the characteristics of neighboring buildings. The height of the left (west) adjacent building, with a 23.07% and complex effect, and building density (22.99%) show an inverse relationship with PV energy generation, as denser configurations and taller adjacent buildings increase shading and reduce solar exposure. The distance to the northeast building, with a 22.5% effect, demonstrates a direct relationship, where reduced distance leads to decreased PV energy generation due to increased shading. The south building distance, at 22.17%, also influences PV energy generation in a complex manner. The adjacent building height to the right (east), at 22.81%, and northwest building distance, with a lesser impact of 2.62%, influence PV energy generation, with greater distances generally enhancing solar access. The distances to the west, east, and north buildings follow in the magnitude ranking of influential variables, with dW showing an inverse relationship and both dE and dN having complex effects on PV energy generation.

Several negligible variables were identified during statistical significance testing. These include the distance to the southeast, green space ratio, WWR, and the heights of the west and northwest buildings, all of which were excluded from ranking due to their minimal impact on PV energy generation. Additionally, the commercial ratio, building footprint, street width, and the heights of the east and north buildings were found to have the least influence on PV energy generation and were excluded from prioritization.

### 4.2.6. Sky View Factor (SVF)

The dominant variable affecting SVF is street width showing an inverse relationship. As street width increases, plot sizes become smaller, leading to reduced courtyard areas for each building, which in turn limits the SVF. Building shape also plays a critical role, with a 69.53% influence. Rectangular buildings generally provide more sky view compared to L-shaped buildings. Building length, at 55.52%, has a direct relationship with SVF, as longer buildings create larger open spaces, enhancing sky exposure.

Among the influential variables are several related to the neighboring building characteristics and site layout. The height of the right (east) adjacent building, with a 45.22% impact, and height of the left (west) adjacent building, at 44.05%, show direct relationships with SVF, where reducing the height of adjacent buildings increases sky exposure. The number of stories, with a 37.81% impact, demonstrates a complex relationship with SVF, as taller buildings may obstruct the sky view in some cases, while in others, the additional height could enhance it depending on the layout. The southwest building distance, with a 36.82% influence, exhibits an inverse relationship, as greater distance reduces the SVF by reducing sky exposure. Similarly, the west building distance, at 32.57%, shows an inverse relationship, where increasing this distance reduces sky view. Other influential factors include building area (31.84%), density (31.45%), and building footprint (31.33%). Increased building density generally reduces sky exposure, as more crowded configurations obstruct the view of the sky. Site length (31.20%) directly influences the size of courtyard areas, with longer sites allowing for more open space and greater sky view. The distance to the east building, with a 28.66% influence, shows a direct relationship with SVF, where reducing the distance increases sky exposure. Building width (22.27%) has an inverse relationship, where wider buildings reduce sky exposure by occupying more horizontal space.

Some variables exhibit minimal influence and are considered negligible. For instance, the distance to the southeast building, at 18.8%, shows an inverse relationship, suggesting that greater distance might obstruct sky visibility. The south building distance, with a 20.01% effect, and the height of the northwest building, at 13.82%, both exhibit positive effects on SVF when reduced, as lower or closer neighboring buildings allow for greater sky view. Lastly, several other variables were excluded from prioritization during significance testing, including the WWR, building use, and the distance to the northwest building, due to their limited influence on SVF.

## 5. Discussion

Figure 6 displays the normalized SHAP values obtained from the sensitivity analysis exhibiting the magnitude ranking of each variable with respect to energy and environmental metrics.

The analysis identified building shape and WWR as the most influential variables affecting cooling, lighting, and heating energy. Building shape shows an influence of 94.1% on cooling energy, and first influential variable on heating and lighting energy, while WWR has an influence of 80% on cooling energy, 74.6% on heating energy, and 77.7% on lighting energy. These findings align with the research by Shang and Hou [3] and Talami and Jakubiec [20][75], who emphasize the importance of considering building and window geometry during the early design stage to minimize energy demand. Building density emerges as a crucial variable, with the greatest impact on the sunlight hours received on facades (45 %), followed by its effect on cooling energy (35.8%). This result is consistent with the work of Yu et al. [26], who identified density as a key factor in the energy performance of residential communities. Similarly, Natanian et al. [33] highlighted density and orientation as key inputs in urban studies, reflecting the substantial influence of density on energy dynamics.

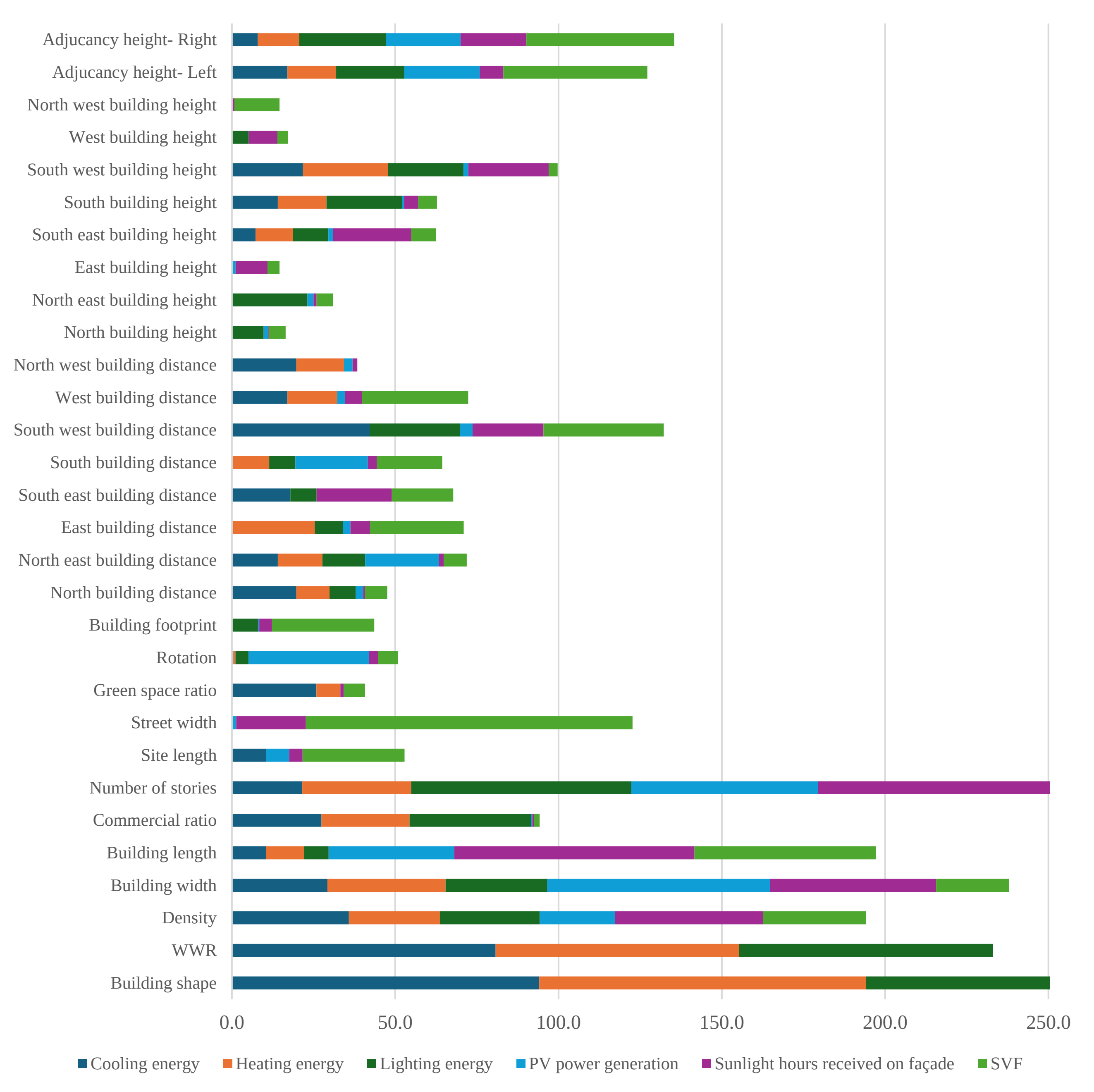

Figure 6. Normalized SHAP magnitude values between variables and energy and environmental metrics.

Following in order of importance are building length and building width. Building length significantly affects the sunlight hours received on facades (73.5%) and displays a notable impact on PV power generation with 58.5% and SVF (55.5%), while building width significantly affects

PV power generation (68.3%) and sunlight hours received on facades (50.8%). These findings are supported by research in Singapore that identified building footprint as the most important input in urban morphology, significantly influencing solar performance [76].

The number of stories displayed a notable impact on all metrics, with the greatest effect on sunlight hours received on facades, followed by lighting energy (67.4%) and PV power generation (57.2%). This observation is supported by Cajot and Schüler [77], who highlighted building height as a critical determinant of energy requirements. These findings are in line with Ahmadian et al.[78], who demonstrated that compact, low-rise forms with larger plan depths improve energy efficiency in temperate climates. In the dry-arid climate of Tehran, building density similarly impacts cooling and heating energy, though additional factors such as solar exposure and PV generation play a more prominent role in this analysis.The analysis also showed that the commercial ratio is an influential variable for energy metrics, affecting lighting energy by 37.1% and cooling energy by 27.4%, although it exhibited a less significant relationship with other output metrics, such as PV power generation (0.43%) and SVF (1.7%). This nuanced role in energy performance reflects the complexity of building usage. Street width affects SVF (100%) and sunlight hours received on facades (21.3%). This is supported by research showing that street geometry and orientation influence the amount of solar radiation received by street surfaces, affecting thermal comfort and building energy consumption [7][24].

Site length emerged as the most significant factor for SVF (31.2%), while the green space ratio, impacts cooling energy (25.9%) and heating energy (7.3%). These results are consistent with findings that the urban layout, including building footprint and green spaces, influences solar performance in open spaces [79]. Rotation affects all output metrics, with the most significant impact on PV power generation (56.8%). The importance of orientation in determining energy outcomes is echoed by Cajot and Schüler [77], who identified orientation as a primary factor in energy requirement analysis. In conclusion, the analysis of building energy and environmental metrics is well-supported by existing literature, highlighting the importance of design variables such as shape, orientation, density, and footprint in influencing energy dynamics and solar performance in urban environments.

Regarding the characteristics of neighboring buildings, the heights and distances of surrounding buildings had varying impacts on each output metric. Overall, it can be concluded that the most influential variable is the southwest building distance, which showed the greatest impact on cooling energy (42.2%), followed by SVF (36.8%). Next is the southwest building height (hSW), with the greatest impact on heating (26%), cooling (21.8%) and lighting energy (23.1%), and the least impact on PV power generation (1.5%). Subsequently, the heights of the buildings adjacent to the left (west) and right (east) displayed the largest impact on SVF (44.1% and 45.2%, respectively), energy metrics, and PV power generation. Overall, geometric characteristics of each building, such as width, shape, density, number of floors, and building length, can be regarded as the most influential variables for most output metrics. Variables like building footprint, and the heights of buildings located to the north, west, northwest, and northeast are considered the least influential. Figure 7 summarizes the influence of each variable on the investigated energy and environmental metrics, categorizing them into three levels: dominant, influential, and negligible.

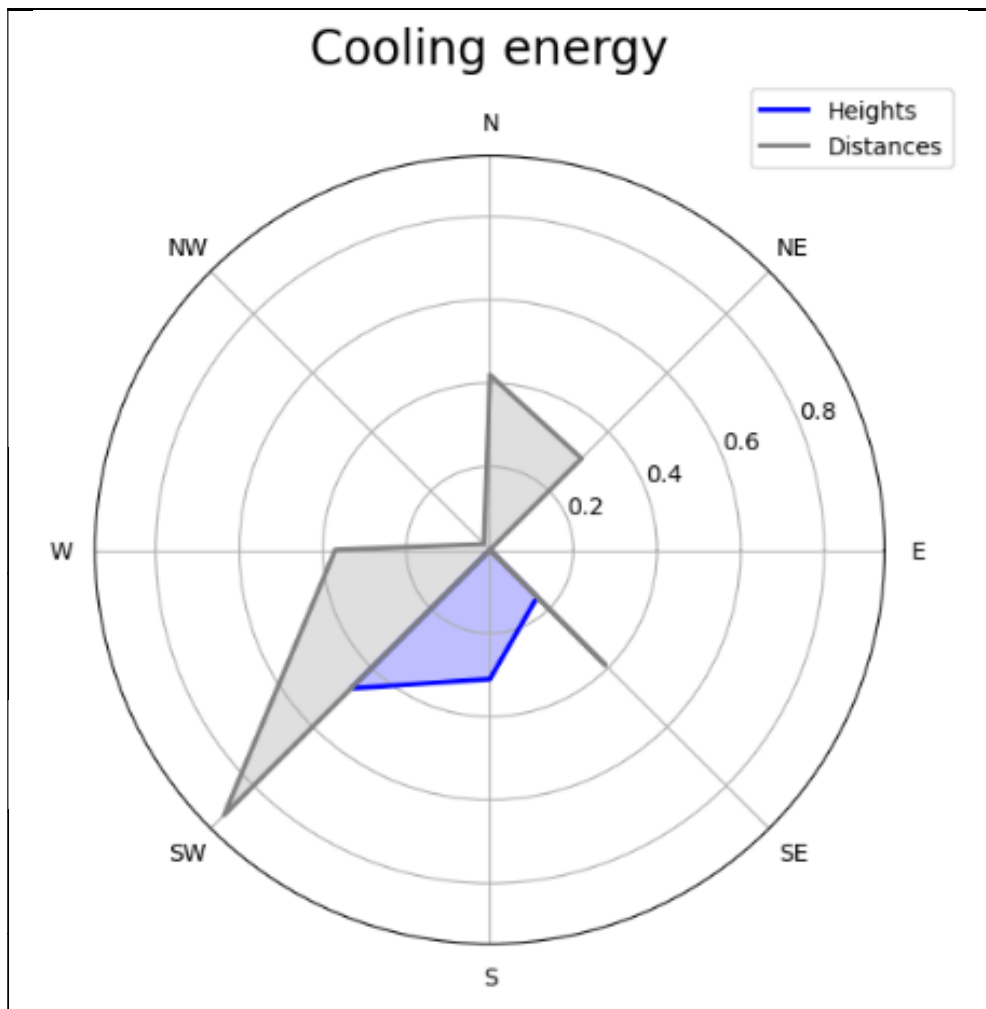

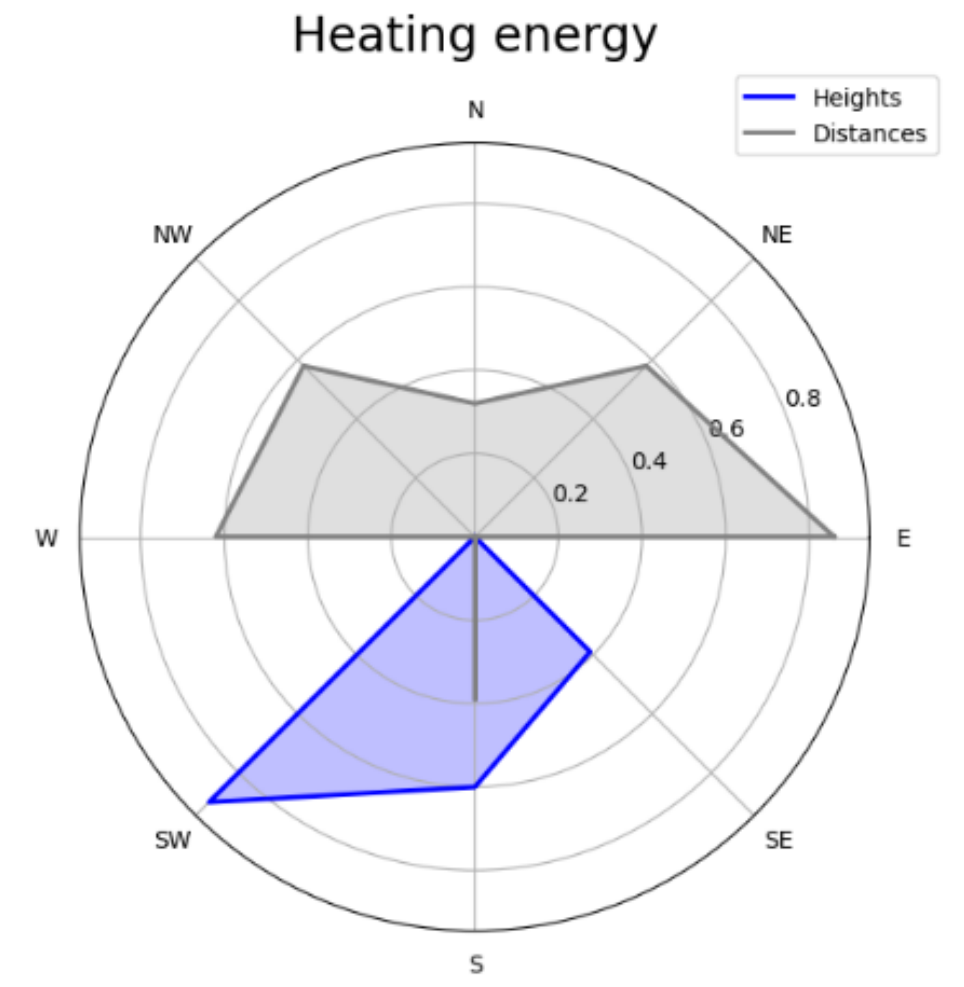

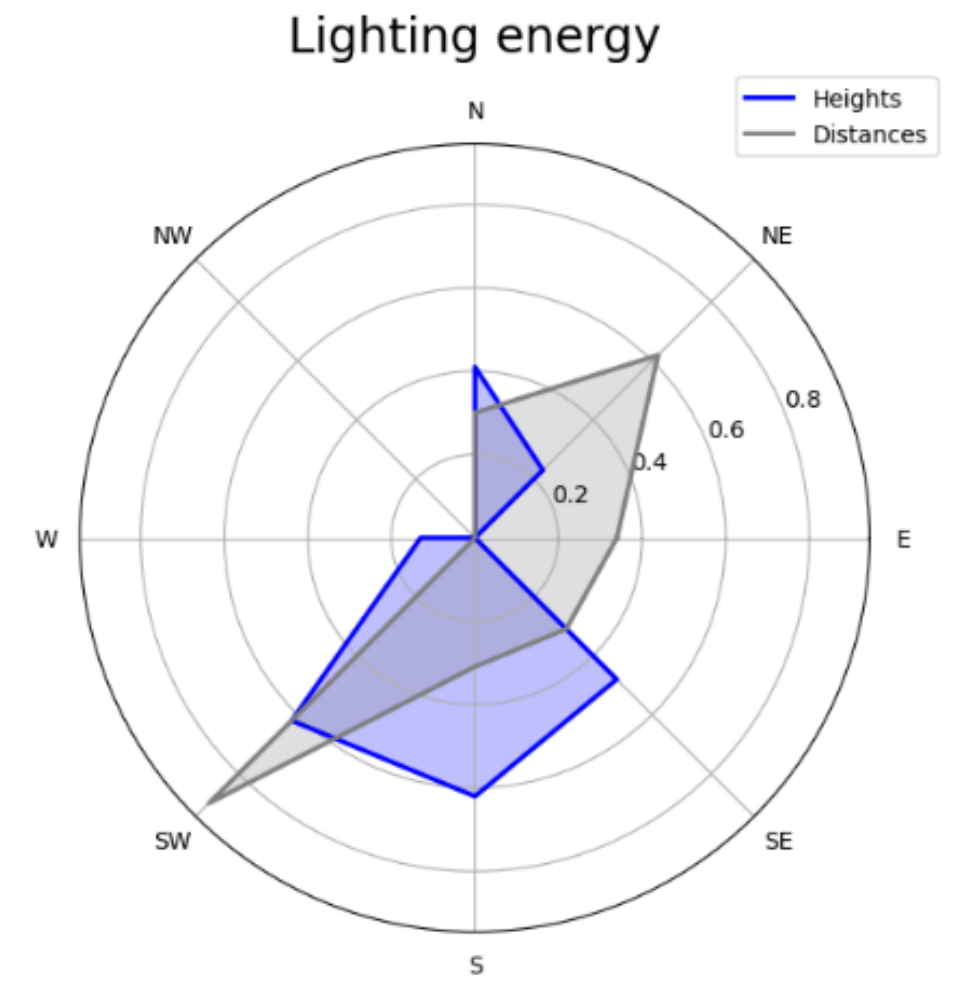

| **Lighting energy** | **Heating energy** | **Cooling energy** | **Classification ranking** |
|---|---|---|---|
| Building shape | Building shape | Commercial ratio | Dominant |
| Commercial ratio | Commercial ratio | Building shape | |
| WWR | WWR | WWR | |
| Number of stories | - | - | |
| - | | | |
| Building width | Building width | Southwest building distance | Influential |
| Density | Number of stories | Density | |
| Southwest building distance | Southwest building height | Building width | |
| Right adjacent building height | Density | Southwest building height | |
| Northeast building distance | East building distance | Green space ratio | |
| South building height | Left adjacent building height | Southeast building distance | Negligible |
| Left adjacent building height | South building height | West building distance | |
| Southeast building height | Rotation | Left adjacent building height | |
| North building height | Northwest building distance | Site length | |
| East building distance | Northeast building distance | Northeast building distance | |

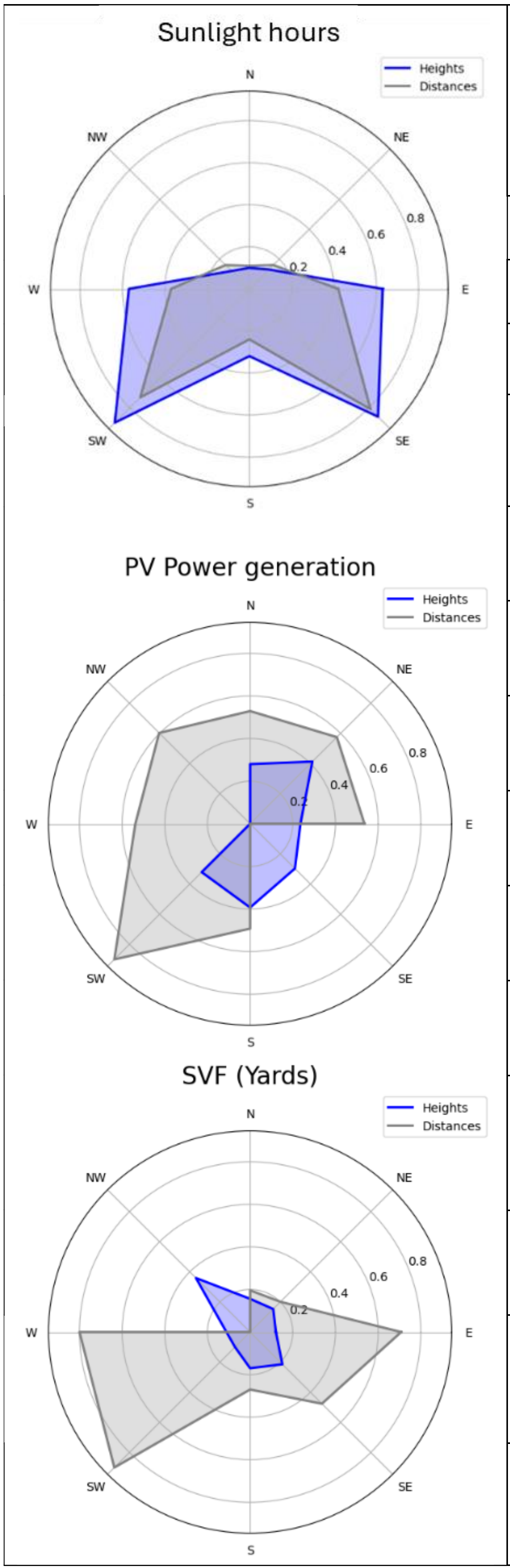

| SVF | PV power generation | Sunlight hours received on facades | Classification ranking |
|---|---|---|---|
| Street width | Building shape | Number of stories | Dominant |
| Building shape | Building width | Building length | |
| Building length | Building length | Building width | |
| - | Rotation | - | |
| | Number of stories | | |
| Southwest building distance | Northeast building distance | Density | Influential |
| West building distance | South building distance | Southwest building height | |
| Density | Left adjacent building height | Southeast building height | |
| Site length | Density | Southwest building distance | |
| Southwest building distance | Right adjacent building height | Southeast building distance | |
| Northwest building distance | Northwest building height | East building height | Negligible |
| Building footprint | Northwest building distance | Building shape | |
| South building distance | West building distance | Street width | |
| Building width | East building distance | West building height | |
| Rotation | North building distance | Left adjacent building height | |

Figure 7. Magnitude ranking of variables for each energy and environmental metric.

## 6. Conclusion

This study provides a detailed exploration of the impact of urban morphology on energy consumption and environmental performance at the urban block scale, with a focus on Tehran's dry-arid climate. By analyzing cooling, heating, and lighting energy demand, as well as PV power generation, sunlight hours on façades, and SVF, the research offers insights into optimizing urban form for greater energy efficiency. The research methodology, which involved parametric and predictive modeling, using various MLMs, and SHAP analysis for feature importance assessment, allowed for a detailed understanding of the variables influencing energy consumption and environmental metrics. The results indicate that while RF and XGBoost algorithms show high predictive accuracy, XGBoost offers a more efficient performance in terms of training time.

Key findings of sensitivity analysis highlight the critical role of building shape, WWR, width, height, and commercial ratio in determining energy performance, particularly for cooling, heating, and lighting energy. Building density also emerged as a critical variable, particularly for sunlight exposure and cooling energy demand. The study also underscores the importance of neighboring buildings' characteristics, such as their height and distance, which significantly affect energy metrics, solar access, and SVF. For instance, While distances generally have a greater influence on cooling energy, both distances and heights are important for heating and lighting energy. PV power generation and SVF are more sensitive to building distances, highlighting the need for careful spatial planning in urban design. While for sunlight hours received on facades, although distances and heights are both important, heights tend to have a more substantial impact, indicating that urban planners must carefully consider the spatial arrangement of buildings to optimize energy efficiency across multiple performance metrics.

Overall, this research contributes to the broader field of UBEM by providing valuable quantitative data and insights specific to the urban block scale within a dry-arid climate. The findings underscore the importance of integrating energy, environmental, and social considerations into urban planning to achieve sustainable development.

Future research should explore how these relationships vary across different climates and urban contexts, expanding the applicability of the findings. Additionally, investigating the integration of emerging technologies such as other renewable energy systems and real-time urban energy monitoring could further enhance energy performance in future urban designs. Lastly, an interdisciplinary approach that integrates architectural design, urban planning, and environmental science is essential to developing holistic solutions that address the full range of sustainability challenges. This study's limitations include its reliance on predefined variable values for variables like building dimensions, street widths, and typologies, which were derived from Tehran's master plan. Furthermore, simplifications in modeling environmental metrics, such as SVF and PV power generation, may not fully capture real-world complexities like vegetation, air pollution, or material reflectivity. These factors highlight the need for further research to assess the broader applicability of the findings and improve accuracy in more variable urban contexts. Nonetheless, the research makes a significant contribution to understanding urban morphology's role in energy and environmental performance, particularly in dry-arid climates.

### Declaration of competing interest

The authors have no competing interests to declare that are relevant to the content of this article.